\pdfoutput=1

\documentclass[11pt]{article}

\usepackage{acl}

\usepackage{times}
\usepackage{latexsym}

\usepackage[T1]{fontenc}

\usepackage[utf8]{inputenc}

\usepackage{microtype}

\usepackage{enumitem}
\usepackage{tablefootnote}

\usepackage{gb4e}
\noautomath

\usepackage{booktabs}

\usepackage{caption}
\usepackage{graphicx, subfig}

\title{Probing for Understanding of English Verb Classes and Alternations in Large Pre-trained Language Models}

\author{David K.~Yi, James V.~Bruno, Jiayu Han, Peter Zukerman, Shane Steinert-Threlkeld \\
Department of Linguistics, University of Washington\\
  \texttt{\{davidyi6, jbruno, jyhan126, pzuk, shanest\}@uw.edu}}

\begin{document}
\maketitle
\begin{abstract}
We investigate the extent to which verb alternation classes, as described by \citet{levin1993}, are encoded in the embeddings of Large Pre-trained Language Models (PLMs) such as BERT, RoBERTa, ELECTRA, and DeBERTa using selectively constructed diagnostic classifiers for word and sentence-level prediction tasks. We follow and expand upon the experiments of \citet{kann-etal-2019-verb}, which aim to probe whether static embeddings encode frame-selectional properties of verbs. At both the word and sentence level, we find that contextual embeddings from PLMs not only outperform non-contextual embeddings, but achieve astonishingly high accuracies on tasks across most alternation classes. Additionally, we find evidence that the middle-to-upper layers of PLMs achieve better performance on average than the lower layers across all probing tasks.
\end{abstract}

\section{Introduction}
We investigate the extent to which verb alternation classes are represented in word and sentence embeddings produced by Pre-trained Language Model (PLM) embeddings \citep{Qiu_2020}.  As first comprehensively cataloged by \citet{levin1993}, verbs pattern together into classes according to the syntactic alternations in which they can and cannot participate.  For example, (\ref{ex:good-caus-inch}) illustrates the \textit{causative-inchoative} alternation.  \emph{Break} can be a transitive verb in which the subject of the sentence is the agent and the direct object is the theme, as in example (1a).  It can also alternate with the form in (1b), in which the subject of the sentence is the theme and the agent is unexpressed.
However, (\ref{ex:bad-caus-inch}) demonstrates that \emph{cut} cannot participate in the same alternation, despite its semantic similarity.

\begin{exe}
    \ex
        \label{ex:good-caus-inch}
        \begin{xlist}
            \ex[] {Janet broke the cup.}
            \ex[] {The cup broke.}
        \end{xlist}

    \ex
        \label{ex:bad-caus-inch}
        \begin{xlist}
            \ex[]{Margaret cut the bread.}
            \ex[*]{The bread cut.}
        \end{xlist}
\end{exe}

(\ref{ex:good-spray-load}) demonstrates an alternation of a different class -- namely, the \emph{spray-load} class, in which the theme and locative arguments can be syntactically realized as either direct objects or objects of the preposition.  \emph{Spray} can participate in the alternation, but as shown in (\ref{ex:bad-spray-load}), \emph{pour} cannot.

\begin{exe}
    \ex 
        \label{ex:good-spray-load}
        \begin{xlist}
            \ex[] {Jack sprayed paint on the wall.}
            \ex[] {Jack sprayed the wall with paint.}
        \end{xlist}

    \ex 
        \label{ex:bad-spray-load}
        \begin{xlist}
            \ex[] {Tamara poured water into the bowl.}
            \ex[*] {Tamara poured the bowl with water.}
        \end{xlist}
\end{exe}

The alternations in which a verb may participate is taken to be a lexical property of the verb \citep[e.g.][]{pinker1989,levin1993,levin1995unaccusativity,shafer2009causative}.  Moreover, we hypothesize that the alternations should be observable within large text corpora, and are therefore available during the pre-training procedure for PLMs such as BERT \citep{bertpaper}.  In contrast, ungrammatical examples such as (2b) and (4b) should be virtually absent from the training data. This leads us to hypothesize that PLM embeddings should encode whether particular verbs are allowed to participate in syntactic frames of various alternation classes.  Our research questions are as follows:

\begin{enumerate}
    \item Do PLM word-level representations encode information about which syntactic frames an individual verb can participate in?
    \item At the sentence level, do PLM embeddings encode the frame-selectional properties of their main verb?
    
\end{enumerate}

Through our series of experiments, we find that PLM embeddings indeed encode information about verb alternation classes at both the word and sentence level. While performance is relatively consistent on the word-level task for the four PLMs we analyze, we find that ELECTRA \citep{clark2020electra} significantly outperforms the other models for the sentence-level task. Furthermore, we find evidence suggesting that middle-to-upper layers encode more information about verb alternation classes since they consistently improve upon the lower layers across all tasks.

The rest of the paper is organized as follows: after a brief review of related literature in Section~\ref{sec:litreview}, we present datasets and models that are relevant to our experiment in Sections~\ref{sec:data} and \ref{sec:models}. We then present two experiments to answer our research questions in Sections~\ref{sec:experiment1} and \ref{sec:experiment2}.  Section~\ref{sec:control-task} presents an additional \textit{control task} \citep{hewitt2019designing} to test whether our linear probes are selective for the given tasks. Finally, we offer a discussion in Section~\ref{sec:discussion} and overall conclusions in Section~\ref{sec:conclusion}.

\section{Related work}
\label{sec:litreview}
Our work follows \citet{kann-etal-2019-verb}, who attempt to predict verb class membership and sentence grammaticality judgments on the basis of GloVe embeddings \citep{glove} and embeddings derived from the 100M-token British National Corpus with a single-directional LSTM \citep{warstadt2019neural}. For the sentence-level task, they further process the input embeddings using a sentence encoder trained on a ``real/fake'' sentence classification task. Varying multi-layer perceptron (MLP) architectures are used for the classification step. Because their primary research focus has to do with how neural language models inform learnability (in the sense of human language acquisition), they intentionally use smaller language models derived from ``an amount of data similar to what humans are exposed to during language acquisition'' and avoid models trained on ``several orders of magnitude more data than humans see in a lifetime'' (p. 291). 

As described in Section \ref{sec:experiment1}, we depart and build upon \citealt{kann-etal-2019-verb} by examining the embedding representations of PLMs instead of static embeddings. We then use an intentionally simple and selective linear diagnostic classifier to probe the representations, as our research questions focuses on the PLM embeddings themselves. We note that \citet{kann-etal-2019-verb} achieves only modest performance in prediction accuracy and MCC, and only for a limited number of verb classes. While this is a valuable result for their research goals, our hypothesis is that PLMs will achieve better performance due to a combination of their contextual representations, complex architectures, and larger training corpora.

To our knowledge, attempting to predict verb alternation class membership along the lines of \citealt{levin1993} from PLM representations is novel. However, two very closely related lines of work include the experiments of \citet{https://doi.org/10.48550/arxiv.1901.03438}, which respectively evaluate the performance of various PLMs on the CoLA \citep{warstadt2019neural} and BliMP \citep{warstadt-etal-2020-blimp} benchmarks, which include acceptability judgment examples from a wide variety of linguistic phenomena (including verb argument structures). We distinguish our experiments from these papers in two major ways. First, we attempt to directly probe the linguistic knowledge of individual PLM embedding layers with a classification probe instead of specifically finetuning the models to a specific task. Second, we limit our focus to verb alternation classes and present detailed analysis about patterns and trends across different alternations and their corresponding syntactic frames. 

\begin{table*}[ht]
\small
\centering
\begin{tabular}{lrrrrrrrrrr}
\toprule
\textsc{levin-class} & \multicolumn{2}{l}{\textsc{caus-inch}} & 
\multicolumn{2}{l}{\textsc{dative}} &
\multicolumn{2}{l}{\textsc{spray-load}} & \multicolumn{2}{l}{there-\textsc{insertion}} & \multicolumn{2}{l}{\textsc{understood-object}} \\
{} &  Inch. &  Caus. &  Prep. &  2-Obj &  with &  loc. &  no-there &  there &  Refl &  No-Refl \\
\midrule
Positive &     73 &    124 &     65 &     74 &   101 &    86 &       149 &     50 &    84 &       11 \\
Negative &    144 &      0 &    377 &    442 &   242 &   257 &         0 &    192 &   419 &      503 \\
\midrule
Total    &    217 &    124 &    442 &    516 &   343 &   343 &       149 &    242 &   503 &      514 \\
\bottomrule
\end{tabular}
\caption{An updated overview of the LaVA dataset based on verb membership class distributions for each syntactic frame. ``Postitive'' refers to the number of verbs that can participate in the specified syntactic frame, while ``Negative'' refers to the number of verbs that cannot participate.} 
\label{table:lava}
\end{table*}

\section{Data}
\label{sec:data}
In our experiments, we use two dataset created by \citet{kann-etal-2019-verb}. One is the \textbf{L}exic\textbf{a}l \textbf{V}erb-frame \textbf{A}lternations dataset (LaVA), which is based on the verbs and alternation classes defined in \citet{levin1993}. It contains a mapping of $516$ verbs to $5$ alternation classes, which are further subdivided into two syntactic frames for each alternation. The broad categories of the alternation classes are: \emph{Spray-Load}, \emph{Causative-Inchoative}, \emph{Dative}, \emph{There-insertion}, and \emph{Understood-object}. Table \ref{table:lava}\footnote{A similar table appears in \citet{kann-etal-2019-verb}, but we present it again here because of discrepancies that we found in the distribution counts. Notably, it appears that the authors flipped the positive and negative counts for the \emph{there-Insertion} and \emph{Understood-Object} alternation classes which carries over to their results. \label{footnote:lava}} provides the class distributions for each syntactic frame. \textbf{F}rames and \textbf{A}lternations of \textbf{V}erbs (FAVA), the other dataset, is a corpus of 9413 semi-automatically generated sentences formed from the verbs in LaVA along with human grammaticality judgments. The sentences in FAVA are categorized according to the relevant alternation class, and are separated into train, development, and test sets by the authors for each category.

\section{Models}
\label{sec:models}

In addition to BERT, we perform experiments on several recent Transformer-based PLMs including RoBERTa \citep{liu2019roberta}, DeBERTa \citep{he2021deberta}, and ELECTRA \citep{clark2020electra} which vary from BERT in a few ways including modifications to BERT's tokenization and pre-training procedure and the size of their training corpus. To make comparisons between each model fair, we use the base architectures for each model which have 12 layers, 12 attention heads, and a hidden layer size of 768.\footnote{All further references to these models refer to their \textit{base} architectures.} 

\subsection{Model differences}
For pre-training, BERT uses standard Masked Language Modeling (MLM) wherein tokens from a given input sequence are masked at random and the model attempts to recover the masked tokens from the unmasked tokens and Next Sentence Prediction (NSP), in which the model tries to predict whether one sentence follows another in a given text sequence. The other PLMs drop NSP from their pre-training procedure but make other significant changes to the architecture and the MLM approach. RoBERTa introduces ``dynamic'' masking, in which different tokens are masked across different training epochs (as opposed to the same training mask being used across epochs). DeBERTa uses a ``disentangled attention mechanism'' which computes attention weights using distinctly encoded position and context vectors, and also moves absolute position encodings from the input layer to the second-to-last layer. Lastly, instead of randomly masking input tokens, ELECTRA strategically replaces tokens with plausible alternatives using a trained generator network, and separately trains a discriminative model which aims to predict whether each token in an input sequence was replaced by a generator sample. 

\subsection{Training Data}
In addition to variations in the pre-training methods, the models are also trained on different datasets. BERT and ELECTRA are both trained on the English Wikipedia Dump and BookCorpus \citep{Zhu_2015_ICCV}. DeBERTa is additionally trained on CC-Stories \citep{https://doi.org/10.48550/arxiv.1806.02847} and OpenWebText \citep{Gokaslan2019OpenWeb}. Finally, RoBERTa is pretrained on all of the aforementioned datasets as well as the CC-News corpus \citep{Mackenzie2020CCNewsEnAL}. 

\section{Experiment 1: Frame Membership from Word Embeddings}
\label{sec:experiment1}

\subsection{Method}

\begin{table*}[ht]
\small
\centering
\tabcolsep=0.11cm
\hspace*{-0.6cm}\begin{tabular}{lrrrrrrrrrrr}
\toprule
{} & \multicolumn{5}{c}{MCC} & \multicolumn{5}{c}{Accuracy} \\ \cmidrule(r){2-6} \cmidrule(r){7-11}
{} &   \multicolumn{1}{c}{Ref.} &  \multicolumn{1}{c}{BERT} & \multicolumn{1}{c}{DeBERTa} & \multicolumn{1}{c}{ELECTRA} & \multicolumn{1}{c}{RoBERTa} &
\multicolumn{1}{c}{Ref.} &  \multicolumn{1}{c}{BERT} & \multicolumn{1}{c}{DeBERTa} & \multicolumn{1}{c}{ELECTRA} & \multicolumn{1}{c}{RoBERTa}  \\
\midrule
\textsc{Causative-Inchoative} & & & & & & & & & & \\
\hspace{1em} Inchoative   &    0.555 & 0.948 [11] & 0.969 [11] & 0.959 [5] & 0.969 [7] & 0.855 &  0.977 & 0.986 & 0.982 & 0.986 \\
\hspace{1em} Causative \textsuperscript{*} &  0.000 & 0.000 & 0.000 & 0.000 & 0.000 & 1.000 &  1.000 & 1.000 & 1.000 & 1.000 \\
\midrule
\textsc{Dative} & & & & & & & & & &       \\
\hspace{1em} Preposition       &    0.320 & \textbf{0.954 [8]} & 0.937 [12] & 0.945 [11] & 0.928 [9] & 0.850 &  \textbf{0.989} & 0.984 & 0.986 & 0.982 \\
\hspace{1em} Double-Object     &    0.482 & 0.976 [10] & 0.968 [10] & 0.976 [12] & 0.936 [9] &   0.853 &  0.994 & 0.992 & 0.994 & 0.984 \\
\midrule
\textsc{Spray-Load} & & & & & & & & & &   \\
\hspace{1em} With              &    0.645 & 0.972 [10] & 0.972 [12] & \textbf{0.979 [8]} & 0.930 [10] & 0.839 &  0.988 & 0.988 & \textbf{0.991} & 0.971 \\
\hspace{1em} Locative          &    0.253 & \textbf{0.969 [10]} & 0.961 [12] & 0.961 [9] & 0.953 [11] & 0.734 &  \textbf{0.989} & 0.985 & 0.985 & 0.983 \\
\midrule
\textsc{There} & & & & & & & & & &    \\
\hspace{1em} No-There \textsuperscript{*}        &    0.000 & 0.000 & 0.000 & 0.000 & 0.000 & 1.000 &  1.000 & 1.000 & 1.000 & 1.000 \\
\hspace{1em} There             &    0.459 & 1.000 [9] & 0.987 [7] & 1.000 [10] & 0.962 [10] &  0.858 &  1.000 & 1.00 & 1.000 & 0.988 \\
\midrule
\textsc{Understood Object } & & & & & & & & & &      \\
\hspace{1em} Refl    &    0.000 & 0.868 [6] & 0.869[12] & 0.860 [5] & \textbf{0.884 [11]} & 0.732 &  0.964 & 0.964 & 0.962 & \textbf{0.968} \\
\hspace{1em} Non-Refl        &    0.219 & 0.850 [7] & 0.850 [10] & \textbf{0.855 [11]} & 0.794 [8] &   0.976 &  0.994 & 0.994 & 0.994 & 0.992 \\
\bottomrule
\end{tabular}
\caption{Results from Word-Level experiments with static embeddings. Reference MCC is from  \protect\citet{kann-etal-2019-verb}'s CoLA word-level experiments. The \textsuperscript{*} symbol indicates syntactic frames which only have positive examples, which trivially achieve 100\% accuracy and 0 MCC (see Footnote~\ref{footnote:lava}). The best performing model for a given frame is denoted in \textbf{bold} (ties are not bolded), and the best performing layer for each model is denoted in brackets `[]'.}
\label{tab:contextual-word-results}
\end{table*}

In order to answer the first question: ``Do PLM token-level representations encode information about which syntactic frames an individual verb can participate in?'', we build a diagnostic classifier for each syntactic frame which takes a verb's layer embedding representation as input.  For example, to probe the \textit{Spray-Load} alternation, we build two binary classifiers: one that predicts whether a verb can participate in the ``locative'' frame and one that predicts whether a verb can participate in the ``with'' frame.  

Furthermore, we build a separate classifier for each model layer based on the embedding representations from that particular layer. For the token-embedding layer, the verb embedding is formed by averaging the pretrained token embeddings that correspond to a particular verb. For layers 1--12, the verb embedding is formed by incorporating contextual information from the sentences in FAVA. Specifically, for each verb, we pass the grammatical sentences from FAVA that contain the verb as input to the PLM and average over the token embeddings corresponding to the verb. We then average over the verb representations for all input sentences in each layer to form the ``layer-embedding'' for the verb. 

We choose a Logistic Regression classifier without regularization as our diagnostic probe as implemented in \texttt{scikit-learn} \citep{sklearn} and show that it is sufficiently \textit{selective} in Section \ref{sec:control-task}. Following \citet{kann-etal-2019-verb}, we use stratified k-fold cross-validation to split the verbs into 4 equally-sized folds: 3 of which are chosen to be the training set and the remaining fold chosen to be the test set.

Also following \citet{kann-etal-2019-verb}, we report Matthews correlation coefficient (MCC) \citep{mcc} in addition to accuracy for model evaluation. MCC is better suited to data such as ours, in which there is an extreme majority class bias for all syntactic frames.
\footnote{All code and data needed to replicate our analysis can be found at \url{https://github.com/kvah/analyzing_verb_alternations_plms}} 

\subsection{Results}

\begin{figure}[htbp]
  \centering
  \setlength{\abovecaptionskip}{0.1cm}
  \includegraphics[width=8cm]{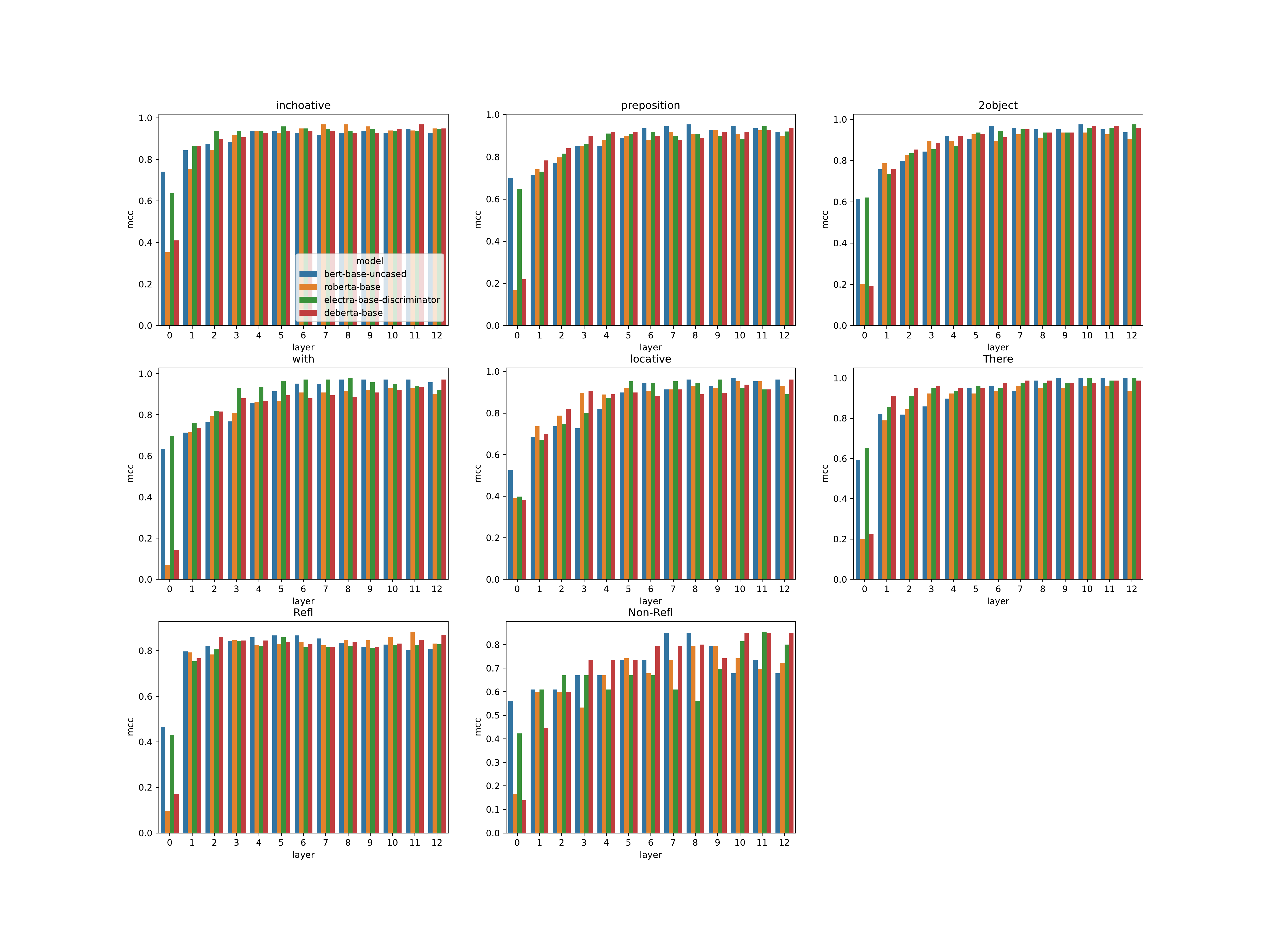} 
  \caption{MCC for each model layer across all syntactic frames on LAVA}
  \label{fig:word-results}
\end{figure}

In Figure~\ref{fig:word-results}, we present the layer-by-layer performance of each PLM and in Table~\ref{tab:contextual-word-results}, we report a comparison between the best-performing layer for each PLM alongside the performance of the ``CoLA-style'' reference embeddings from \citet{kann-etal-2019-verb}. Overall, we find that the contextual PLM embeddings dramatically outperform the reference embeddings in terms of both MCC and accuracy.

\begin{figure}[htbp]
  \centering
  \setlength{\abovecaptionskip}{0.1cm}
  \includegraphics[width=7.5cm]{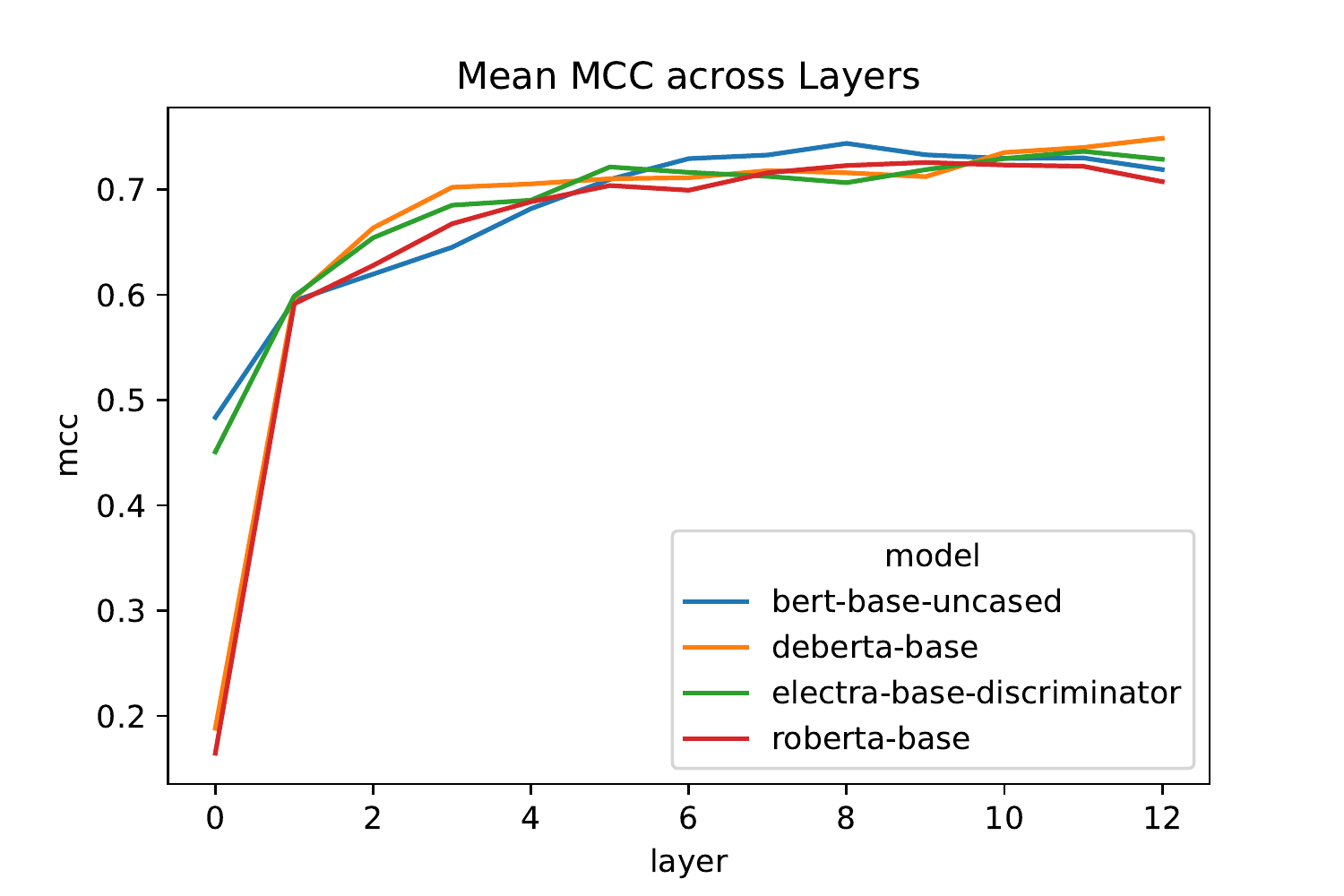} 
  \caption{Mean MCC for each model layer across all syntactic frames on LAVA}
  \label{fig:mean-results-word}
\end{figure}

Surprisingly, the PLMs perform well even for the more challenging frames; for the ``locative'' frame, BERT achieves 0.969 MCC compared to 0.253 when using the reference embeddings, and for the ``non-reflexive'' frame, ELECTRA achieves 0.855 MCC compared to 0.219 when using the reference embeddings. Furthermore, we observe consistent patterns in performance across different layers of each PLM. As shown in Figure~\ref{fig:mean-results-word}, the lower layers achieve low-to-moderate correlation on average while the middle-to-upper layers consistently achieve strong correlation.

\section{Control Task}
\label{sec:control-task}
A control task as described by \citep{hewitt2019designing} aims to combat the \textit{Probe Confounder Problem}, which highlights the issue of supervised probe classifiers ``learning'' a linguistic task by combining signals in the data that are irrelevant to the linguistic property of interest. In the context our first experiment, a confounding probe would be problematic since it suggests that good model performance may be attributed to arbitrary signals picked up by the probe, as opposed to the PLM embeddings actually containing linguistic information about the syntactic frames. To mitigate the Probe Confounder Problem, we implement an example control task for the \textit{Spray-Load} ``with'' syntactic frame for BERT. 

For each verb $v_i$ in LaVA with a binary label $y_i$ denoting whether $v_i$ can participate in the syntax frame \textsc{sl-with}, we independently sample a control behavior $C(v)$ by randomly assigning a binary ``membership'' value to $v_i$ based on the empirical membership distribution of verbs that participate in the \textsc{sl-with} syntax frame. The control task is the function that maps each verb, $v_i$, to the label specified by the control behavior $C(V_i)$:

$$f_{control}(v_i) = C(v_i)$$

By construction, the control task should only be learnable by supervised probe. Following the experiment design of \citet{hewitt2019designing}, we compare the \textit{selectivity} of a linear probe, an Multi-layer Perceptron with 1-hidden layer (MLP-1), and an MLP with 2-hidden layers (MLP-2) where the \textit{selectivity} of a model is defined by the difference between its accuracy on the real task (i.e. predicting verb membership for the \textsc{sl-with} frame) and the control task. In addition, we explore several ``complexity control'' methods including limitation of feature dimensionality, reducing the number of training examples, and increasing regularization. 

\subsection{Complexity Hyperparameters}

In this section, we describe the complexity control methods in more detail and enumerate the hyperparameters that we tried for each method. The control parameters were chosen based on the three most effective methods from the experiments of \citet{hewitt2019designing}. To isolate the effect of each control method, we only change one of the complexity parameters in each experiment.

\subsubsection{Limiting Dimensionality}

For the Logistic Regression model, we reduce the dimensionality of the feature embeddings by performing a Truncated Singular Value Decomposition and limiting the output matrix to rank $k$. For the MLP models, we simply limit the size of the hidden layer(s) to $k$.

Considering the input BERT embeddings which have 768 dimensions, we limit $k$ to the following values: $\{20, 100, 300, 500\}$.

\subsubsection{Reducing Proportion of Training Data}

Because LaVA is not split into train and test sets, we use 4-fold cross validation as done in \citet{kann-etal-2019-verb} with 3 training folds and one test fold for evaluation on the control task. As an additional constraint, we reduce the number of training samples in each training fold by randomly sampling a proportion $p$ of the samples and discarding the rest. 

Although \citet{zhang-bowman-2018-language} recommend training on 1\%, 10\% and, 100\% of the training data, our training data is relatively small and imbalanced (71\% of the train set verbs do not participate in the \textsc{sl-with} frame). Hence, we experiment with larger values of $p$: $\{0.1, 0.3, 0.5, 0.7, 0.9\}$

\subsubsection{$L_2$ Regularization}

For both the linear and MLP models, we add $L_2$ regularization with the following strength values: $\{0.01, 0.1, 0.2, 0.5, 1\}$

\subsection{Results}

\begin{figure}[htbp]
  \centering
  \setlength{\abovecaptionskip}{0.1cm}
  \includegraphics[width=8cm]{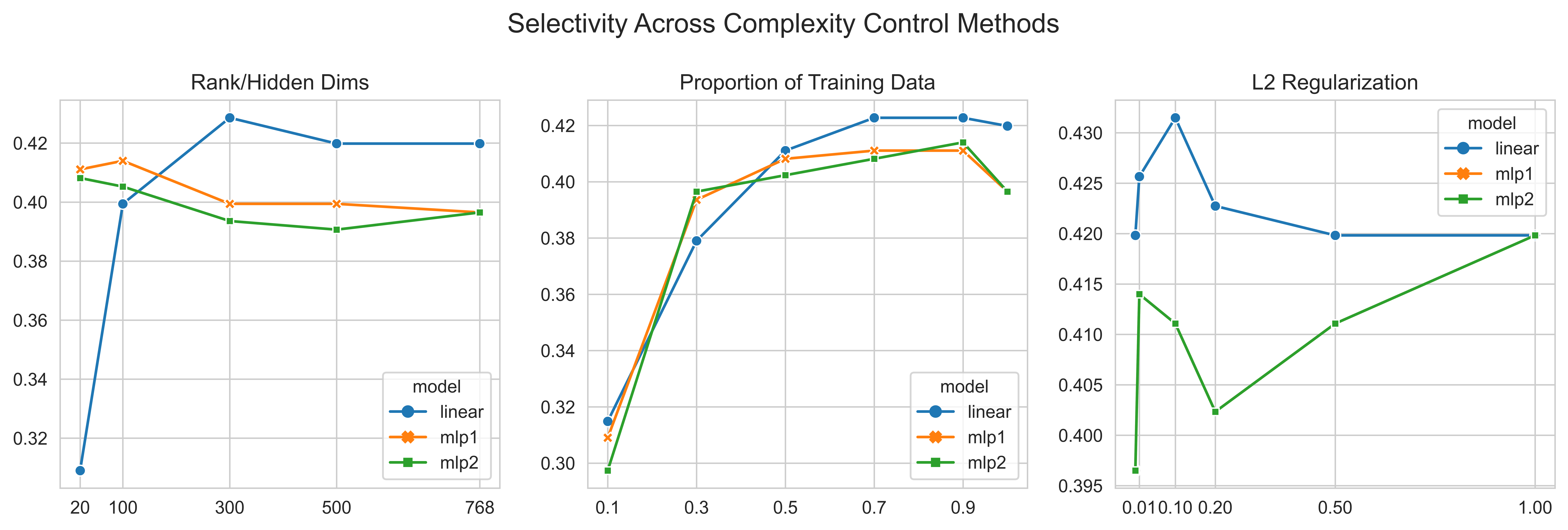} 
  \caption{Linguistic Task selectivities for the three complexity control methods. }
  \label{fig:selectivity}
\end{figure}

Figure~\ref{fig:selectivity} shows the high-level trends across experiment configurations for model selectivity. We observe that the linear model with 
default parameters ($k=768, p=1, L_2=0$) outperforms both the MLP-1 and MLP-2 model in selectivity (0.420 v.s. 0.397) with no significant decrease in linguistic task accuracy (0.985 for the linear and MLP-1 models v.s. 0.988 for the MLP-2 model). 

\begin{figure}[htbp]
  \centering
  \setlength{\abovecaptionskip}{0.1cm}
  \includegraphics[width=8cm]{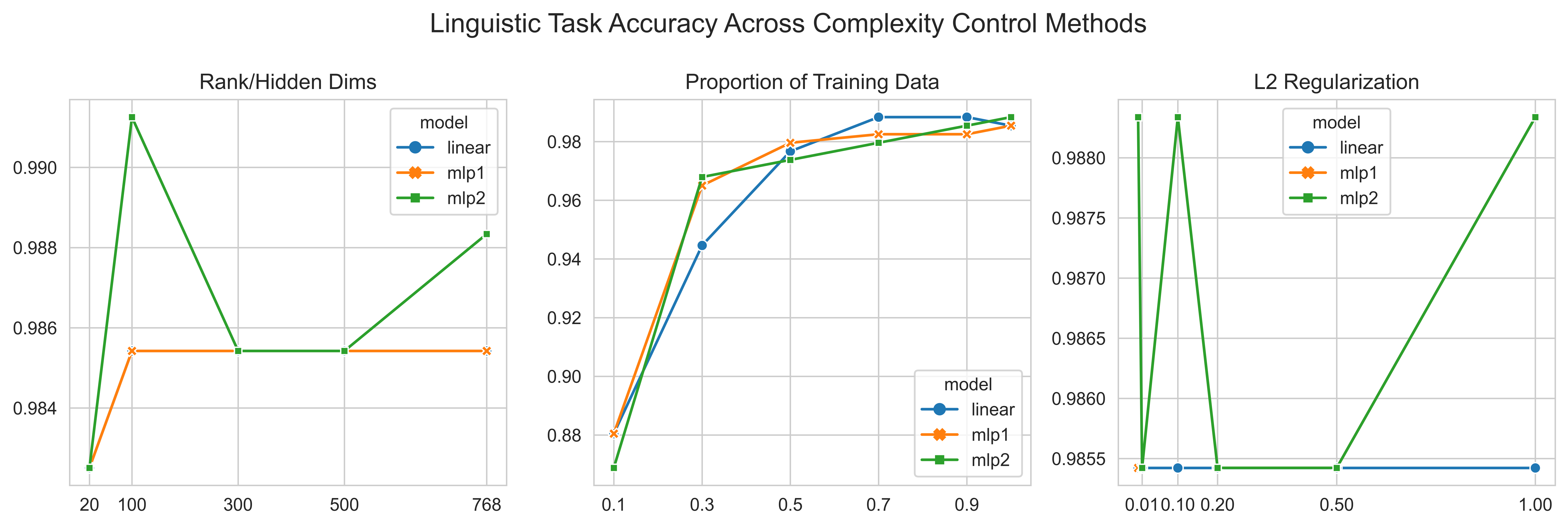} 
  \caption{Linguistic Task accuracies for the three complexity control methods. }
  \label{fig:accuracy}
\end{figure}

Looking at the effect of complexity control methods on model accuracy in Figure~\ref{fig:accuracy}, we find that limiting dimensionality and $L_2$ regularization has little impact across all configurations, with the worst model (linear: $k=20$) achieving an accuracy of 0.983 and the best model (MLP-2: $k=100$) achieving only a slightly higher accuracy of 0.991. On the other hand, reducing the proportion of data in each training fold appears to have significant impact on model performance. For the linear model, there is a huge discrepancy in accuracy between training on 10\% of the data (0.869) and the full training set (0.988). A nearly identical pattern can be observed for both of the MLP models as well.

\vspace{-0.3cm}

Comparing selectivity, the linear models outperform both MLPs across all complexity control methods. For dimensionality control, we see a lower selectivity in the linear model for lower values of $k$ ($k=20, 100$) but the best linear model ($k=300$) achieved a higher selectivity (0.429) than the best MLP model (0.414). Similarly, the best performing configuration for reduced training samples and $L_2$ regularization are linear models with $p=0.9$ (0.423) and $L_2=0.1$ (0.431) respectively. 

\vspace{-0.3cm}

We arrive at two major conclusions from the control task experiments. The first is that a linear probe is a good choice for our linguistic task since it achieves higher selectivity than the MLP models without substantial loss in model accuracy across a wide range of complexity control methods. The second is that limiting dimensionality, reducing training samples, and $L_2$ regularization are all effective methods for increasing model selectivity for both the linear and MLP models. However, the best configurations are not significantly better (> 0.01 improvement in selectivity) than the default linear model so we did not make any modifications to our classification probe. As we only performed these experiments for BERT and the \textsc{sl-with} syntactic frame specifically, a great avenue for future work is to test whether our results extend to other PLMs and syntactic frames. 

\section{Experiment 2: Grammar Judgments from Sentence-embeddings}
\label{sec:experiment2}

\subsection{Method}

In the second experiment, we investigate the extent to which PLMs encode frame-selectional properties of their main verb. For each PLM and embedding layer, we fit a binary Logistic Regression classifier on the FAVA training set for a given alternation class which predicts whether a given sentence is grammatical. We ignore the held out development set because the probe hyperparameters do not need to be tuned and directly evaluate each model on the test set. The whole process can be described by the following equation:
$$c_{{s}_{i}} = f(\textbf{W}\textbf{s}_{i} +\textbf{b})$$
where $\textbf{s}_{i}$ refers to the embedding of the whole sentence for layer $i$ (by averaging all $i$ layer's hidden states of words in the sentence $s$), $f$ refers to the logistic regression classifier, $\textbf{W}$ and $\textbf{b}$ are the parameters of $f$, and $c_{{s}_{i}}$ is a binary value corresponding to whether the sentence is grammatical. We then extract the best performing layer for each model and compare the results with the reference acceptability judgment model proposed by \citet{kann-etal-2019-verb}.

\begin{table*}[ht]
\small
\centering
\begin{tabular}{lcccccc}
\toprule
{} &  \textsc{Combined} &  \textsc{Causative-inchoative} &\textsc{Dative} &  \textsc{Spray-load} &  \textsc{There} & \textsc{Understood}   \\
\midrule
\textsc{Mcc} & & & & & & \\

\hspace{1em}  \textsc{Ref.} & 0.290 & 0.603 & 0.413 & 0.323 & 0.528 & 0.753 \\

\hspace{1em}  \textsc{BERT} & 0.642 (10) & 0.760 (8) & 0.678 (6) & 0.625 (10) & 0.716 (10) & 0.842 (9) \\

\hspace{1em}  \textsc{DeBERTa} & 0.653 (9) & 0.776 (5) & 0.633 (8) & 0.704 (12) & 0.744 (6) & 0.826 (1) \\
 
\hspace{1em}  \textsc{ELECTRA} & \textbf{0.818 (11)} & \textbf{0.864 (11)} & 0.670 (8) & \textbf{0.830 (12)} & \textbf{0.828 (12)} & \textbf{0.869 (11)} \\

\hspace{1em}  \textsc{RoBERTa} & 0.496 (8) & 0.725 (8) & \textbf{0.802 (2)} & 0.470 (5) & 0.725 (11) & 0.793 (1) \\
\midrule

\textsc{Accuracy}&          &           &            &           &    &     \\  
\hspace{1em}  \textsc{Ref.} & 0.646 & 0.854 & 0.760 & 0.662 & 0.729 & 0.874 \\

\hspace{1em}  \textsc{BERT} & 0.840 (10) & 0.920 (8) & 0.880 (6) & 0.820 (10) & 0.890 (10) & 0.921 (9) \\

\hspace{1em}  \textsc{DeBERTa} & 0.847 (9) & 0.924 (5) & 0.902 (8) & 0.858 (12) & 0.912 (6) & 0.909 (1) \\

\hspace{1em}  \textsc{ELECTRA} & \textbf{0.920 (11)} & \textbf{0.954 (11)} & 0.897 (8) & \textbf{0.918 (12)} & \textbf{0.937 (12)} & \textbf{0.934 (11)} \\

\hspace{1em}  \textsc{RoBERTa} & 0.787 (8) & 0.909 (8) & \textbf{0.944 (2)} & 0.747 (5) & 0.899 (11) & 0.893 (1) \\

\bottomrule
\end{tabular}
\caption{Results from Sentence-Level experiments.  \textsc{Ref} refers to the reference probing model in \cite{kann-etal-2019-verb}. Bolded values show the best result for each alternation class. `()' indicates the best performing layer for each model.}
\label{table:sentence-results}
\end{table*}

\subsection{Results}

 \begin{figure}[htbp]
  \centering
\setlength{\abovecaptionskip}{0.1cm}
  \includegraphics[width=8cm]{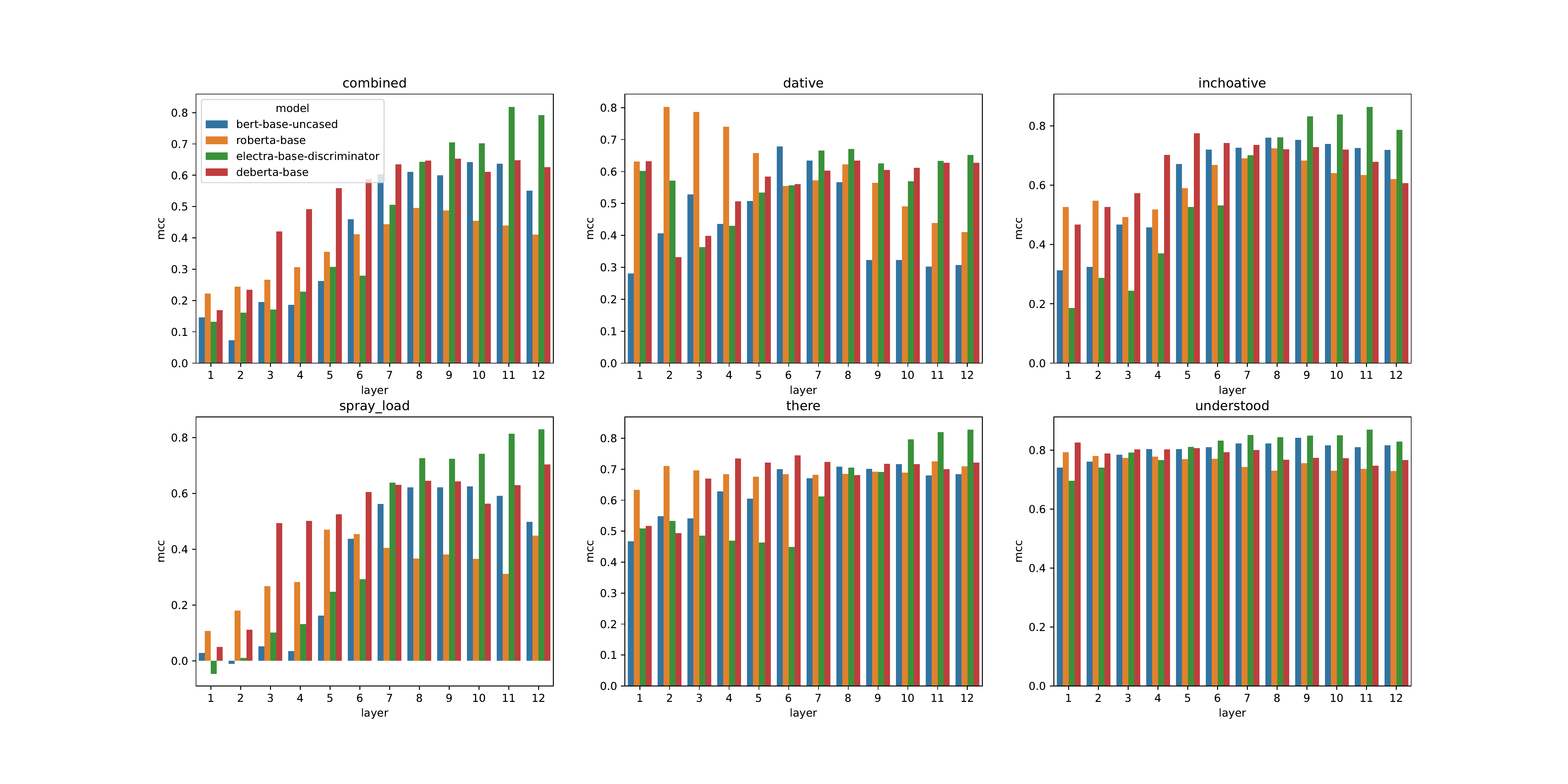} 
  \caption{Layer-by-layer MCC score for each alternation class on FAVA}
  \label{fava:mcc}
\end{figure}

\begin{figure}[htbp]
  \centering
\setlength{\abovecaptionskip}{0.1cm}
  \includegraphics[width=7.5cm]{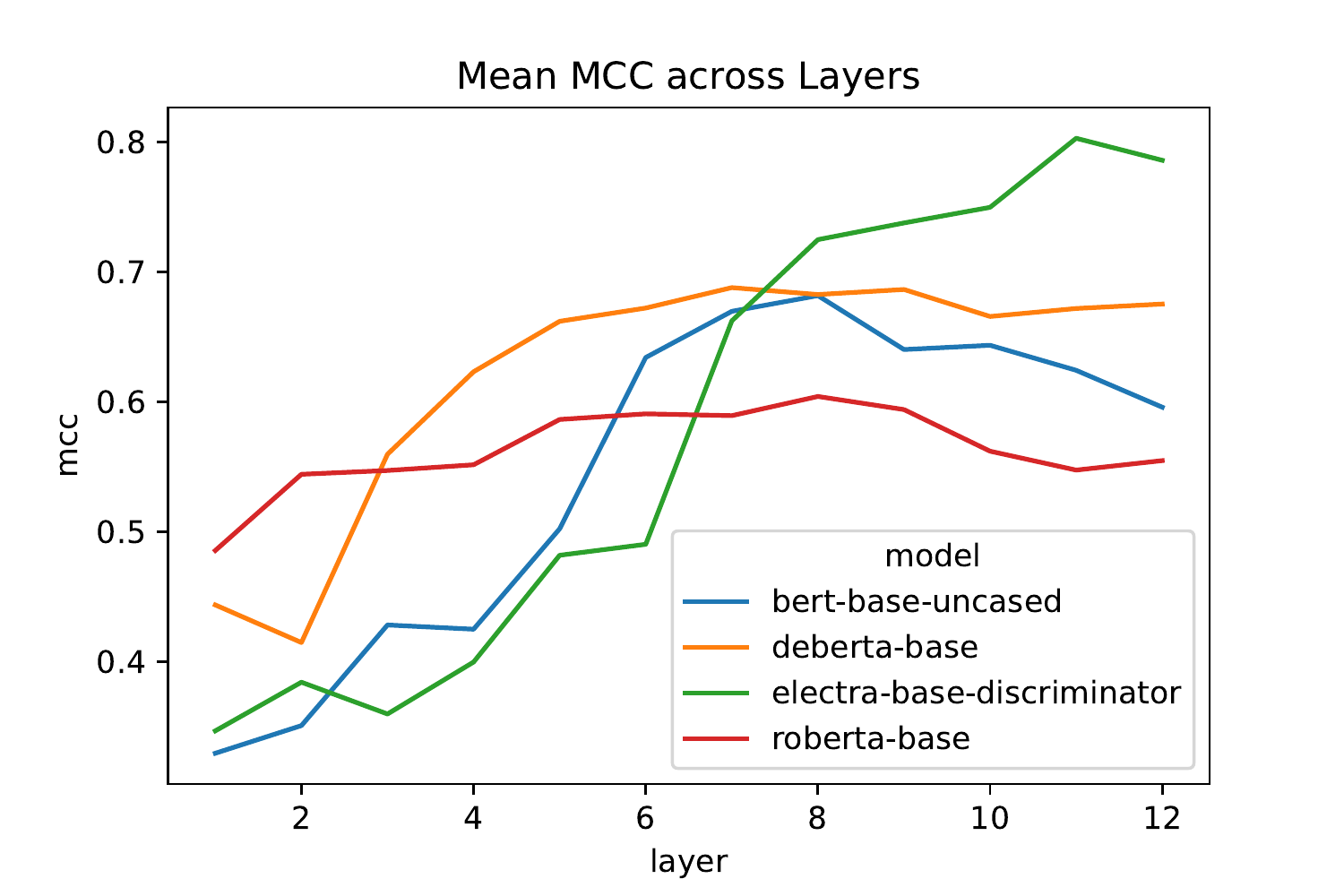} 
  \caption{Mean layer MCC score across all alternation classes on FAVA}
  \label{fig:sentence_trend}
\end{figure}

The MCC and accuracy scores for each model and layer are shown in Figure~\ref{fava:mcc}. From the figure, we can see that there is significant variation in layer performance aside from the \textit{Understood-object} alternation. Generally, we observe a trend in which performance increases substantially from the lower (1-4) layers to the middle layers (5-9), with some models, most notably ELECTRA, continuing to improve through the upper layers (10-12). This can be seen more clearly in Figure~\ref{fig:sentence_trend}, which shows the mean layer performance across each category.
Furthermore, ELECTRA achieves the best MCC on 5 of the 6 categories: \textit{Combined} (0.818), \textit{Inchoative}  (0.864), \textit{Spray\_Load} (0.830), \textit{There} (0.828), and \textit{Understood-Object} (0.869). The outlier frame is RoBERTa, which achieves the best MCC (0.802) on the \textit{Dative} frame. 

Table~\ref{table:sentence-results} provides a comparison between the best performing layer from each PLM and the reference embeddings from \citet{kann-etal-2019-verb} for each alternation class. As defined by \citet{kann-etal-2019-verb} an MCC value between 0.5 and 0.7 demonstrates a moderate correlation between predicted and true labels while an MCC greater than 0.7 implies strong correlation. From the table, we see all models are able to obtain strong correlation for the \textit{Understood-Object} alternation, the \textit{There} frame, and the \textit{Causative-Inchoative} frame. In contrast, BERT and RoBERTa are only able to achieve moderate correlation on the \textit{Spray-Load} frame, while all models except RoBERTa only achieve moderate correlation on the \textit{Dative} alternation. Consistent with the CoLA-style embeddings, we find that the PLMs achieve the best performance on average for sentences from the \textit{Understood-Object} alternation class. This is surprising since frames from the \textit{Understood-Object} alternation were the hardest to predict for the word-level task for both the CoLA-style embeddings and the PLMs. Nevertheless, all PLM outperform the reference model across all alternation categories for the sentence acceptability judgment task.  

\vspace{-0.3cm}

\section{Discussion}
\label{sec:discussion}

On the word level prediction task, all PLMs achieve strong correlation ($>0.7$ MCC) across all syntactic frames with the strongest performance in the ``there'' frame (1.00, achieved by BERT and ELECTRA) and the weakest performance on the ``non-reflexive'' frame (0.794, achieved by RoBERTa). When looking at accuracies, each model is able to predict whether a verb belongs to a particular syntactic frame with excellent accuracy ($>0.95$ across all alternation frames). Morever, looking at Table~\ref{tab:contextual-word-results}, we see that the middle (5-9) and upper (10-12) layers consistently achieve the highest MCC, which is reinforced by the trend shown in Figure~\ref{fig:mean-results-word}. 

For the sentence-level experiments, we see a similar outcome wherein the upper-middle PLM layers achieve the best performance on average. However, we observe that there is much more variation in performance between each PLM. ELECTRA and BERT are relatively consistent, since their best performing layer for all alternation classes either come from the middle or upper layers. In contrast, the lower layers of RoBERTa achieve the best performance on the \textit{Dative} alternation, and both RoBERTa and DeBERTa achieve the best performance on the \textit{Understood-Object} alternation from the first layer. These anomalies can potentially be explained by the claim that different alternation classes require different types of linguistic knowledge (i.e. syntactic v.s. semantic) which are encoded in different PLM layers. However, the consistently strong performance of the upper layers for BERT and ELECTRA across all alternation classes provides counter evidence against the claim. 

ELECTRA is the best performing model overall on the sentence-level acceptability task, achieving the best MCC and accuracy on four of the five alternation classes (all except \textit{Dative}). Unsurprisingly, ELECTRA also excels on the combined dataset compared to the other models (0.165 MCC over the second-best performing model, DeBERTa). While it is difficult to attribute the model’s success to a specific property, one hypothesis is that its generator/discriminator architecture closely resembles the FAVA task of identifying acceptable sentences from linguistic minimal pairs. This idea is reinforced by the authors as well, who note that the model's relatively strong performance on CoLA potentially stems from the fact that the acceptability judgment task of CoLA “closely matches ELECTRA’s pre-training task of identifying fake tokens” \citep[p.15]{clark2020electra}.

While we are optimistic about our results, there are several limitations to our experiments. First, we only analyze five different alternation classes which is a small subset of the 83 classes presented in \citet{levin1993}. In addition -- although our control task ensures that our classifier probe is relatively \textit{selective} for the first experiment and BERT, it may not necessarily generalize well to the second experiment, other syntactic frames, and other models. In the future, we hope to expand our selectivity experiments to a wider array of syntactic frames and models. 
    
\section{Conclusion and Future Work}
\label{sec:conclusion}

Overall, our results support the hypothesis that PLM contextual embeddings encode linguistic information about verb alternation classes at both the word and sentence level. For the frame-selectional verb classification task, all PLMs achieve significant improvement upon the reference CoLA-style embeddings from \citet{kann-etal-2019-verb}, especially for frames in which the CoLA-style embeddings obtain weak correlation (i.e. “locative”, “reflexive”, and “non-reflexive”). Also, it is clear that model performance tends to improve from lower to upper layers, which can be seen the most easily from the mean performance across layer figures. For the sentence acceptability task, we arrive at similar conclusions, albeit with greater distinction in results between different models and layers. While there are numerous factors that may be responsible for the improved performance from PLMs, we hypothesize that the improvement can largely be attributed to the attention-based encodings of transformer models since we only saw modest improvements in performance from the reference embeddings when using the bottom ``static'' layers for each PLM.

In terms of future work, there are several interesting avenues that we hope to explore. From the data perspective, it would certainly be worthwhile to test whether our insights and conclusions extends to the dozens of alternations described in \citet{levin1993} that are not present in the LAVA and FaVA datasets. There are also several interesting adaptations that can be made to our experiment methodology. For example, instead of just analyzing the base architecture for each PLM, we could also analyze \textit{small} and \textit{large} variants to directly evaluate the effect of scaling training data and model size within the same model. Moreover, while we attempt to control the \textit{Probe Confounder Problem} by building a selective probe, there is no guarantee that the classifier probes do not pick up on arbitrary signals in the training data that lead to non-meaningful improvements in performance. Two promising alternative approaches that mitigate this risk include unsupervised evaluation of minimal pairs as shown in \citet{warstadt-etal-2020-blimp} and ``amnesic probing'', which tests whether a property that can be extracted from a probe is actually relevant to task importance \citep{elazar-etal-2021-amnesic}. 

\setcounter{table}{0}
\renewcommand{\thetable}{A\arabic{table}}

\setcounter{figure}{0}
\renewcommand{\thefigure}{A\arabic{figure}}

\bibliography{paper}

\appendix
\onecolumn
\section{Complexity Control Results}
\label{sec:appendix}

\begin{table*}[ht]
\small
\centering
\begin{tabular}{lrrrrr}
\toprule
 {}  &  Dimensions ($k$) &  Training Prop. ($p$) &  $L_2$ Reg. &  Accuracy &  Selectivity \\ 
\midrule
\textsc{Default Params} & & & & & \\
\hspace{1em} Linear &                    768 &                      1.0 &                0.0 &     0.985 &        \textbf{0.420} \\
\hspace{1em} MLP-1 &                    768 &                      1.0 &                0.0 &     0.985 &        0.397 \\
\hspace{1em} MLP-2 &                    768 &                      1.0 &                0.0 &     0.988 &        0.397 \\
\textsc{Limiting Dimensions} & & & & & \\ 
\hspace{1em} Linear &                    300 &                      1.0 &                0.0 &     0.985 &        \textbf{0.429} \\
\hspace{1em} MLP-1 &                    100 &                      1.0 &                0.0 &     0.985 &        0.414 \\
\hspace{1em} MLP-2 &                     20 &                      1.0 &                0.0 &     0.983 &        0.408 \\
 \textsc{Reducing Training Samples} & & & & & \\
\hspace{1em} Linear &                    768 &                      0.9 &                0.0 &     0.988 &        \textbf{0.423} \\
\hspace{1em} MLP-1 &                    768 &                      0.9 &                0.0 &     0.983 &        0.411 \\
\hspace{1em} MLP-2 &                    768 &                      0.9 &                0.0 &     0.985 &        0.414 \\
 \textsc{$L_2$ Regularization} & & & & & \\
\hspace{1em} Linear &                    768 &                      1.0 &                0.1 &     0.985 &        \textbf{0.431} \\
\hspace{1em} MLP-1 &                    768 &                      1.0 &                1.0 &     0.988 &        0.420 \\
\hspace{1em} MLP-2 &                    768 &                      1.0 &                1.0 &     0.988 &        0.420 \\
\bottomrule
\end{tabular}
\caption{Results from the Complexity Control Experiments. For each experiment, only the best performing configuration for each model is reported.}
\label{table:control-results}
\end{table*}

\end{document}